\let\origvec\vec

\documentclass[runningheads]{llncs}

\let\vec\origvec

\usepackage{amsmath}
\usepackage{amssymb,amsfonts}
\usepackage{graphicx}
\usepackage{dsfont}
\usepackage{xcolor}
\usepackage{subfigure}
\usepackage{subfiles}
\usepackage{algpseudocode, algorithm}
\usepackage{array}
\usepackage{multirow}
\usepackage{mathrsfs}
\usepackage{rotating}
\usepackage{pdflscape}

\newcolumntype{C}[1]{>{\centering\arraybackslash}}
\newcolumntype{L}[1]{>{\raggedright\let\newline\\\arraybackslash\hspace{0pt}}m{#1}}

\begin{document}

\title{Sampling Unknown Decision Functions to Build Classifier Copies}

\author{Irene Unceta\inst{1,2}\orcidID{0000-0002-7422-1493} \and
Diego Palacios\inst{1}\orcidID{-} \and
Jordi Nin\inst{3}\orcidID{0000-0002-9659-2762} \and
Oriol Pujol\inst{2}\orcidID{0000-0001-7573-009X}}

\authorrunning{I. Unceta et al.}

\institute{BBVA Data \& Analytics, Barcelona, Spain \\
\email{\{irene.unceta,diego.palacios.contractor\}@bbvadata.com} \and
ESADE Universitat Ramon LLull, Barcelona, Spain \\
\email{jordi.nin@esade.edu} \and
Department of Mathematics and Computer Science, Universitat de Barcelona (UB), Barcelona, Spain \\
\email{oriol\_pujol@ub.edu}}

\maketitle              

\begin{abstract}
Copies have been proposed as a viable alternative to endow machine learning models with properties and features that adapt them to changing needs. A fundamental step of the copying process is generating an unlabelled set of points to explore the decision behavior of the targeted classifier throughout the input space. In this article we propose two sampling strategies to produce such sets. We validate them in six well-known problems and compare them with two standard methods. We evaluate our proposals in terms of both their accuracy performance and their computational cost.

\keywords{Sampling \and Classification \and Copies \and Data generation}
\end{abstract}

\section{Introduction}
\label{sec:intro}

As the use of machine learning models continues to grow and evolve, so does the need to adapt them to a changing context. In many everyday situations, constraints related to production software requirements \cite{Sculley2015HiddenSystems}, external threats  \cite{Fredrikson2015ModelCountermeasures,Shokri2017MembershipModels} or specific regulations \cite{Goodman2017EuropeanExplanation} limit the performance of decision systems. In those cases, an originally functional solution is rendered obsolete by the requirements of its operational environment.

Model re-training may seem a straightforward approach to solve such issues. However, there exist many situations where this procedure is neither advisable nor possible. Take company production environments, for example, where model performance needs to be maintained in time, or scenarios where the training data is no longer available, access to the obsolete model is limited and its internals unknown. In such cases, copies \cite{copying} have been proposed as a suitable alternative to correct performance deviations by substituting the obsolete models with new ones that operate roughly in the same way and yet meet the new requirements.

Copying corresponds to building a machine learning classifier that replicates the decision behavior of another. Works dedicated to knowledge transfer between models have traditionally focused on scenarios where the training data distribution is directly \cite{Hinton2015Distilling} or indirectly~\cite{Caruana2006} known and where rich information outputs can be used as soft targets for the new model \cite{Liu2018Teacher,Yang2018KnowledgeDI}. Copies are envisaged for more restrictive settings: the overall decision behavior needs to be preserved and access to the obsolete model is limited to a membership query interface that outputs only hard predictions. Moreover, all information regarding the training data distribution is assumed to be lost, so that we lack any real data. 

In this article we explore different methods to generate unlabeled data points in the described context. We annotate these points using the trained obsolete model as an oracle and obtain a new labelled dataset to build the copy. In particular, we propose two sampling techniques: an exploration-exploitation policy using a Boundary sampling model and a modified fast Bayesian sampling algorithm that uses Gaussian processes to reduce the uncertainty of the decision function. For comparative purposes, we also use a modified Jacobian sampling strategy based on \cite{Papernot2017PracticalLearning} and uniform sampling. 

In all cases, we allow ourselves no knowledge of model specifics or training data. Instead, we only have access to class predictions. To proof the validity of these techniques we use a set of empirical metrics and conduct experiments using well-established datasets and classifiers. Copies built using the generated data converge to an optimal solution for a sufficiently large number of samples, independently of the sampling technique. While Boundary sampling is better suited to learn linear problems, it is Fast Bayesian sampling which yields the best performance for less points.

The rest of this paper is organized as follows. First we introduce  a literature survey of related work in section~\ref{sec:related}. We then provide an overview of the copying problem in section~\ref{sec:preliminaries}. In section~\ref{sec:algorithms} we present the proposed methods to generate synthetic datasets. section~\ref{sec:experiments} describes the experiments carried out to validate our proposals, together with a set of metrics to evaluate their performance. Section~\ref{sec:discussion} discusses our main findings. Finally, in section~\ref{sec:conclusions} we summarize our main experimental results and outline future research.

\section{Related work}
\label{sec:related}

Drawing samples from arbitrary probability distributions is a well known problem in both mathematics and statistics. In this context, research has been traditionally focused on how to extract a representative data sample from a give population. In machine learning, sampling methods are widely used when training probabilistic models, for tasks such as data augmentation. 

Previous research includes Markov Chain Monte Carlo methods \cite{gilks2,tierney} or sampling of specific probability distributions \cite{NIPS2010_3940}, as well as methods for rejection sampling \cite{dymetman,gilks,pmlr-v5-mansinghka09a} and importance sampling to reduce the variance in Monte Carlo estimation \cite{doucet}. Contrary to traditional sampling methods, however, in the copying context sampling of an original non-probabilistic hard decision function cannot be performed under an implicit probability density function. Instead, we need to choose an initial probability distribution and then incorporate the information we gather along the way to guide future explorations.

In line with this, several works have demonstrated the utility of active learning strategies as opposed to passive learning with random sampling \cite{faster,Dagan95committee-basedsampling}, specially when dealing with imbalanced datasets \cite{imbalanced}. In this scenario a query selection protocol decides whether to accept samples that are passed along to a human annotator \cite{settles.tr09}. Most relevant to this paper are works on uncertainty sampling~\cite{Lewis1994,Lewis1994Heteregeneous} and Bayesian active learning~\cite{Houlsby2011Bayesian}. A key difference between these strategies and ours is that, when selecting the samples to be annotated, most active learning frameworks evaluate the informativeness of unlabeled instances in terms of class probability outputs. In our case, only hard predictions are available. Moreover, while the main objective of active learning is to minimize the cost of annotation, when copying this cost is zero.

A seemingly related, but vastly different problem comes from the area of \textit{transferability}-based adversarial learning  \cite{Papernot2017PracticalLearning,Liu2017DelvingAttacks}, where a malicious adversary exploits samples crafted from a local substitute of a model to compromise it. Here, adversarial attacks focus on a local approximation of the target decision boundary to foil the system. Our aim, however, is to sample the input space to obtain an unlabeled set of points that enables a copy.

\section{Preliminaries}
\label{sec:preliminaries}

Copying \cite{copying} refers to the process of building a machine learning classifier whose decision function resembles that of another, only accessible through a membership query interface and whose internals and training data remain unknown.

\subsection{The copying problem}

Let us take a classifier $f_\mathcal{O}: \mathcal{X} \to \mathcal{T}$, with $\mathcal{X}$ and $\mathcal{T}$ the sample and label spaces, respectively; $\mathscr{D} = \{ (\boldsymbol{x}_i, t_i) \}_{i=1}^M$ the training data and $M$ the total number of  samples. We restrict to case where $\mathcal{X} = \mathbb{R}^d$ and $\mathcal{T} = \mathbb{Z}_k$ for $k$ the number of classes, i.e. classification of real-valued attributes. Copying amounts to building a new classifier $f_\mathcal{C}(\theta)$, the \textit{copy}, parameterized by $\theta$, such that its decision function mimics that of $f_\mathcal{O}$ all over the space. This problem can be understood as finding the optimal parameter values $\theta^*$ that maximize the posterior probability

\begin{equation}
    \theta^* = \arg\max_{\theta} P(\theta|f_{\mathcal{O}})
\label{eq:copy_inference_log_int}
\end{equation}

In practice, maximizing (\ref{eq:copy_inference_log_int}) requires that we sample the input space to gain information about the specific form of $f_\mathcal{O}$. Because the training data $\mathscr{D}$ is unknown we cannot resort to it, nor can we estimate its distribution. In other words, we are left to explore the input space $\mathcal{X}$, with the hard predictions provided by $f_\mathcal{O}$ as only guidance. Hence, we introduce synthetic data points $\boldsymbol{z}_j \in \mathcal{X}$ such that we compute the above equation through the integral

\begin{equation}
    \theta^* = \arg\max_{\theta} \int_{\boldsymbol{z}\sim P_Z} P(\theta|f_{\mathcal{O}}(\boldsymbol{z})) dP_Z
\end{equation}

\noindent
for an arbitrary generating probability distribution $P_Z$, from which the synthetic samples are independently drawn. In using $P_Z$ to model the spatial support of the copy, we ensure the copying process is agnostic to both the specific form of  $f_\mathcal{O}$ and the training data distribution. 

 If we assume an exponential family form for all probability distributions involved in the above expression, we can rewrite it as follows

\begin{flalign}
    \theta^* &= \arg\min_{\theta} \bigg[\int_{\boldsymbol{z}\sim P_Z} \gamma_1\ell_1(f_{\mathcal{C}}(\boldsymbol{z},\theta),f_{\mathcal{O}}(\boldsymbol{z})) dP_Z 
    + \gamma_2\ell_2(\theta,\theta^+) \bigg],
\label{eq:copy_inference_}
\end{flalign}

\noindent
for $\ell_1$ and $\ell_2$ a measure of the disagreement between the two models, and $\theta^+$ our prior knowledge about the copy parameters $\theta$. The solution to (\ref{eq:copy_inference_}) dependent on the specific choice of $P_Z$. Moreover, since we cannot draw infinite samples to explore the space in full, direct computation is not possible. Instead, we use the approximation below 

\begin{flalign}
    (\theta^*,\boldsymbol{Z}^*) &= \arg\min_{\theta,\boldsymbol{z}_j\in \boldsymbol{Z}} \bigg[\frac{1}{N}\sum_{j=1}^N \gamma_1\ell_1(f_{\mathcal{C}}(\boldsymbol{z}_j,\theta),f_{\mathcal{O}}(\boldsymbol{z}_j))
    + \gamma_2\ell_2(\theta,\theta^+) \bigg].\label{eq:copy_inference_approx_}
\end{flalign}

\noindent
for $\boldsymbol{Z}^* = \{ \boldsymbol{z}_j \}_{j=1}^N$ an optimal set of synthetic samples. This set is labelled using class prediction outputs of $f_\mathcal{O}$ as hard targets. We refer to the resulting labelled dataset $\mathscr{Z} = \{ (\boldsymbol{z}_j, f_\mathcal{O}(\boldsymbol{z}_j)) \}_{j=1}^N$ as the synthetic dataset.

The form of (\ref{eq:copy_inference_approx_}) resembles that of regularized empirical risk minimization \cite{Vapnik2000}. In contrast to that framework, however, the expression above can be understood as a dual optimization problem, where we simultaneously optimize the model parameters $\theta$ and the set $\boldsymbol{Z}$ used to explore the input space. In the simplest approach, we can cast this problem into one where we only use a single iteration of an alternating projection optimization scheme: the \textit{single-pass copy} \cite{copying}. Under this approach, we split the dual optimization in two independent sub-problems. We first find an optimal set of unlabelled data points $\boldsymbol{Z}^*$ and then optimize for the copy parameters $\theta^*$\footnote{The resulting optimization problem differs from the standard learning process. Copying problems are always separable and, given enough capacity, zero training error is always achievable without hindering the generalization performance of the copy. Although standard machine learning methods work in this context, we argue that ad hoc techniques can be used to exploit these differences when building a copy \cite{copying}}. In this article we address the first part of this optimization to answer the question: {\it which set of synthetic points minimizes the error when substituting the integral in (\ref{eq:copy_inference_}) with a finite sum over $N$ elements?}

\section{Methods}
\label{sec:algorithms}

With the aim of answering this question, we explore different approaches to generating unlabeled data under the mentioned constrains. In particular, we discuss Boundary sampling and Fast Bayesian sampling. 

\subsection{Boundary sampling}

This method combines uniform exploration with a certain amount of exploitation. The main idea is to conduct a targeted exploration of the space until the decision boundary is found. The area around the boundary is then exploited by alternatively sampling points at both sides. Because different decision regions are to be expected, this process is repeated several times to ensure a proper coverage of the whole decision space. The detailed algorithm is shown in Alg. \ref{algo:boundary}. 

We begin by generating samples uniformly at random, until we find a sample whose predicted class label differs from the others (lines 3 to 8). We then proceed to do a binary search in the line that connects the last two samples. This binary search is stopped when a pair of points $(\boldsymbol{z}_a, \boldsymbol{z}_b)$ is found such that $\|\boldsymbol{z}_a-\boldsymbol{z}_b\|_2<\varepsilon$ with $f_\mathcal{O}(\boldsymbol{z}_a)\neq f_\mathcal{O}(\boldsymbol{z}_b)$ for a given tolerance $\varepsilon$, \textit{i.e.} points located at a distance from the boundary no larger than $\varepsilon$ (lines 9 to 13).
Taking one of these two points $\boldsymbol{z}$ as a starting point, we generate a unitary random vector to select the preferred direction towards where the next samples are drawn. We draw samples at a constant step distance $\lambda$ until we find a point $\boldsymbol{z}'$ that belongs to a different class, $f_\mathcal{O} (\boldsymbol{z}) \neq f_\mathcal{O} (\boldsymbol{z}')$, and repeat the process (lines 17 to 25). 

\begin{algorithm}[!t]
\caption{Boundary Sampling(\textbf{int} $N$, \textbf{Classifier} $f_\mathcal{O}$)}
\label{algo:boundary}
\begin{algorithmic}[1]
\State $Z \gets \emptyset$

\While{$|Z| < N$}
    \State $z_a \sim \textit{Uniform}(\mathcal{X}), \quad y_a \gets f_\mathcal{O}(z_a)$ 
    \Repeat \Comment{Search samples with different labels}
        \State $z_b, y_b \gets z_a, y_a$
        \State $z_a \sim \textit{Uniform}(\mathcal{X}), \quad y_a \gets f_\mathcal{O}(z_a)$
        \State $Z \gets Z \cup \{z_a, y_a)\}$
    \Until{$y_a \neq y_b$}
    
    \While{$||z_b - z_a||_2 \geq \varepsilon$}\Comment{binary search}

        \State $z_c \gets (z_a + z_b)/2, \quad y_c \gets f_\mathcal{O}(z_c)$
        \State $Z \gets Z \cup \{(z_c, y_c)\}$
        
        
        \State $z_b, y_b \gets z_c, y_c$ \textbf{if} $y_c \neq y_a$ \textbf{else} $z_a, y_a \gets z_c, y_c$
        
    \EndWhile

    \State $T \gets \{ (z_c, y_c): \text{repeated }\mathcal{I} \text{ times}\}$ 
    \While{$T \neq \emptyset \text{ and for no more than } \mathcal{T}$ iterations}
    \State $(z, y) \in T, \quad T \gets T\setminus \{(z, y)\}$
    \State $ u \gets u/\|u\|_2, \; u \sim \mathbb{N}(0,I_d)$ \Comment{random direction}
    \For{$\mathcal{N}$ times}
        \For{$\alpha \in \{1, 0.9, \cdots, -0.9, -1\} $ and while $f_\mathcal{O}(z + \lambda v) = y$}
            \State $v \gets \alpha \cdot u + w$ with $w$ a random vector s.t. $w\perp u, ||v||_2 = 1$
        \EndFor
        \State $z \gets x + \lambda v,\quad y \gets f_\mathcal{O}(z)$
        \State $Z \gets Z \cup \{(z, y)\},\quad  u \gets v$
        \State every Poiss($\lambda '$) points do $T \gets T\cup (z, y)$
    \EndFor

    \EndWhile
    
\EndWhile

\State \Return S

\end{algorithmic}
\end{algorithm}

For the boundary exploration, the value of $\varepsilon$ must be small compared to the boundary exploration step $\lambda$, which determines the Euclidean distance between two consecutive samples. The lower the value of $\varepsilon$, the closer the initial sample will be. The number of samples in the binary search increases with the logarithm of $1/\varepsilon$. For the case of parameter $\lambda$, the higher its value, the faster the boundary will be covered with less samples, albeit with less resolution. If $\lambda$ is small a large proportion of the boundary may remain unexplored.

The above process results in a set of samples that alternate the two sides of the decision boundary, with distance to the boundary bounded by $\lambda$. These samples define a 1-dimensional curve: a thread. A thread can only contain a predefined number of steps $\mathcal{N}$, which at any given time depends on the number of generated samples. Threads are stopped when out of range or when no other samples are found in the given direction. To ensure a good coverage of the space, we allow $\mathcal{T}$ threads to be created from each point $\boldsymbol{z}$. When this number is reached, a new binary search starts. For high values of $\mathcal{T}$, the boundary is well sampled in certain areas, but many regions are left uncovered. For lower values, more binary searches are performed and consequently more regions of the boundary are explored, with less intensity.

We allow each thread to generate other threads with a frequency modelled by a Poisson distribution parameterized by $\lambda'$. For higher values of $\lambda'$, exploration threads are more dispersed and samples in the vicinity of the boundary more spread. For smaller values, samples are not as spread, but specific regions are more exploited. We perform $\mathcal{I}$ independent runs, increasing the maximum number of threads from run to run. Given a desired number of samples $N$ and to ensure a balanced representation of the space, we generate half following the Boundary sampling algorithm and the other half using random sampling. The theoretical computational cost of this method is $\mathcal{O}(N d)$.

\subsection{Fast Bayesian sampling}
\label{sec:bayesiansampling}

In this method, the function to optimize is assumed to be a random process and samples are generated maximizing an acquisition function. We start assigning a large uncertainty to the whole input space and reduce it everytime a new sample is generated. The goal is to reduce the global uncertainty by guiding future sampling towards the most uncertain areas. 

Let us define $g \sim \mathcal{GP}(0, k_{SE})$ as a Gaussian Process with mean $0$ and a squared exponential kernel of the form

\begin{equation*}
    k_{SE}(\boldsymbol{z}, \boldsymbol{z}') = \sigma^2 e^{-\frac{||\boldsymbol{z} - \boldsymbol{z}'||_2^2}{2l^2}},
\end{equation*}

\noindent
for length scale $l$ and variance $\sigma^2$. Every realization $g_i$ of the stochastic process $g$ is such that $g_i : \mathcal{X} \to \mathbb{R}$. In particular, we treat $f_{\mathcal{O}}$ as one of such realizations\footnote{There exist realizations of $g$ as close to $f_\mathcal{O}$ as desired. }. Our objective is to find a set of points $\boldsymbol{Z}$ such that the function $\lfloor \mathbb{E}[g^{\boldsymbol{Z}}] \rceil$~\footnote{$\lfloor x \rceil$ rounds $x$ to the nearest integer.}, where $g^{\boldsymbol{Z}} = (g \, | \, g(\boldsymbol{z}) = f_\mathcal{O}(\boldsymbol{z}) , \forall \, \boldsymbol{z} \in \boldsymbol{Z})$, is similar enough to $f_\mathcal{O}$. With this aim in mind, we propose an acquisition function ${f: \mathcal{GP} \times \mathcal{X} \to \mathbb{R}}$ of the form

\begin{equation*}
f(g, \boldsymbol{z}) = \mathbb{V}_\textit{ar}[g(\boldsymbol{z})] \cdot [1 + \tau\cdot \textit{frac}(\mathbb{E}[g(\boldsymbol{z})])^2(1 - \textit{frac}(\mathbb{E}[g(\boldsymbol{z})]))^2],
\end{equation*}

\noindent
where $\textit{frac}(\boldsymbol{z})$ stands for the decimal part of $\boldsymbol{z}$. The first term above involves the variance of the process. When choosing the next sample, this term gives a higher priority to points with a bigger variance, \textit{i.e.} located in a region where we have little information about $f_\mathcal{O}$. The second term refines the areas close to the integer numbers which are valid labels, where there is a transition between classes. This provides better knowledge of the function near the boundary. Parameter $\tau$ governs the trade-off between the two terms. 

As it is, finding the \textit{a posteriori} distribution of the Gaussian Process has a high computational cost due to the estimation of the mean and the covariance matrix. Moreover, this cost is increased when optimizing for the maximum of the acquisition function. The total cost is roughly $\mathcal{O}(dN^3)$. To overcome this limitation, we propose a faster version, where we find the \textit{a posteriori} Gaussian distribution in a single optimization, by limiting the number of samples used to compute the \textit{a posteriori} process to $b$. The lower the value of $b$, the faster the algorithm converges, but also the less accurate it is. While this approach is notably faster, it is not warranted to find an optimal solution.

\begin{algorithm}[!t]
\caption{Fast Bayesian Sampling(\textbf{int} $N$, \textbf{Classifier} $f_\mathcal{O}$)}
\label{algo:bay2}
\begin{algorithmic}[1]
\State $Z \gets \{(z, f_\mathcal{O}(z)) \;|\; 10 \times z \sim \textit{Uniform} (\mathcal{D}) \}$
\While{$|Z| < N$}
    \If{$|Z| \leq b$}
        \State $Z_r \gets Z$
    \Else \Comment{Limit to $b$ the number of samples to calculate the posterior distribution}
        \State $Z_r \subseteq Z \; s.t. \; |Z_r| = b$
    \EndIf
    \State $g^{Z_r} \gets g \:|\: g(z) = y \, \forall (z, y) \in Z_r$ \Comment{A posteriori Gaussian process}
    
    \State $T \gets \emptyset$
    \Repeat
        \State $z_0 \sim \textit{Uniform}(\mathcal{D})$
        \State $z \gets \text{argmax}_{z \in \mathcal{V}(z_0) \subseteq \mathcal{D}} \; f(g^{Z_r}, z)$
        \State $y \gets f_\mathcal{O}(z)$
        \State $T \gets T \cup (z, y)$
    \Until {$|T| = \lfloor|Z_r|/\text{sf}\rceil$} \Comment{sf: slowness factor}
    \State $Z \gets Z \cup T$
\EndWhile

\State \Return $S$

\end{algorithmic}
\end{algorithm}

When computing the maximum of the acquisition function, the starting point is set to $z_0 \sim \textit{Uniform}(\mathcal{X})$ and the total number of iterations to $\mathcal{N}$, the number of random samples used to compute the first a posteriori distribution. Given the no convexity (in general) of the acquisition function, we can identify the next point to sample to be $\boldsymbol{z} = \text{argmax}_{\boldsymbol{z} \in \mathcal{V}(\boldsymbol{z}_0) \subseteq \mathcal{D}} \; f(g^{\boldsymbol{Z}}, \boldsymbol{z})$ for $\mathcal{V}(\boldsymbol{z}_0)$ a neighbourhood of $\boldsymbol{z}_0$ not necessarily open. The number of points generated by this Gaussian process is limited by slowness factor $sf$, the inverse of the fraction of samples generated from the gaussian process with respect to those used to compute the \textit{a posteriori}, without recalculation. This factor exploits the fact that the optimization of the acquisition function only finds local minima, i.e. it does not generate the same samples. A low value makes the algorithm faster, but less precise. 

These modifications allow us to sample the acquisition function without having to constantly recalculate the posterior. In all cases, the number of points calculated without reoptimizing the Gaussian process is proportional to the number of samples used to optimize the previous Gaussian process. The full algorithm for Fast Bayesian sampling is depicted in Alg. \ref{algo:bay2}. This method has linear complexity with respect to the number of samples, roughly $\mathcal{O}(Ndb^2)$. Unless otherwise specified, we use the term Bayesian sampling to refer to this faster version.

\section{Experiments}
\label{sec:experiments}

We evaluate these sampling algorithms on 6 UCI datasets. In what follows, we describe the experimental set up and the metrics we use to validate our results. 

\subsection{Experimental settings}
\label{sec:experiments.setting}

We use datasets \textit{bank}, \textit{ilpd-indian-liver}, \textit{magic}, \textit{miniboone}, \textit{seeds} and \textit{synthetic-control}, a hetereogeneous sample of binary and multiclass problems for a real feature space. We assume data are normally distributed and apply a linear transformation such that each attribute has mean $0.5$ and standard deviation $1/5.152$. As a result, when variables are not correlated, $0.99^d$ samples lay inside the $[0,1]^d$ hypercube. We split data into stratified 80/20 training and test sets and fit artificial neural networks (ANNs) with a single hidden layer of 5 neurons.

We generate synthetic sets of size $10^6$ in the restricted the input space $[0,1]^d$. For comparative purposes, we also generate samples using random sampling and an adapted version of the Jacobian-based Dataset Augmentation algorithm, proposed by \cite{Papernot2017PracticalLearning}. The choice of parameters for each method is specified in Table \ref{tab:parameters}. Because all algorithms produce new samples in an accumulative way, we  generate smaller sets by selecting the first $j$ points. Finally, we generate balanced reference sample sets $\mathcal{W} = \{\boldsymbol{w}_i, f_{\mathcal{O}}(\boldsymbol{w}_i) \}_{i=1}^L$, comprised of $L=10^7$ data points sampled uniformly at random in the $[0,1]^d$ hypercube. 

\subsection{Evaluation metrics}
\label{sec:metrics}

We evaluate sampling strategies in terms of the performance of copies built on the synthetic data resulting from each method. For this purpose, we build copies using an ANN with an equivalent architecture as above (ANN), a logistic regression (LR), a decision tree classifier (DT) and a deeper ANN with 3 hidden layers with 50 neurons each (ANN2). 

\begin{table}[!t]
    \renewcommand{\arraystretch}{1.3}
    \caption{Parameters settings.}
    \label{tab:parameters}
    \centering
    \begin{tabular}{L{2.5cm}||L{8.5cm}}
    \hline
        \textbf{Algorithm} & \textbf{Parameters} \\ 
        \hline
        \hline
        \multirow{2}{*}{\textit{Boundary}} & $\varepsilon = 0.01$, $\lambda = 0.05$, $\lambda' = 5$, $\mathcal{I}=\text{round}(2 + \log(N))$ \\ & $\mathcal{T}=\text{round}(8 + 4\log(N))$,  $\mathcal{N}=5 + 2.6\log(N)$\\
        \hline
        \textit{Bayesian} & $l = 0.5 \sqrt{d}$, $\sigma^2 = 0.25 k^2$, $\tau = 10$, $sf = 20$ $b = 1000$, $\mathcal{N} = 10$\\
        \hline
        \textit{Jacobian} & $\mathcal{I} = \min(100, \text{round}(5 + N/4)$, $\mathcal{T} = 50$, $\lambda = 0.05$, $\mathcal{P} = 5$ \\
    \hline
    \end{tabular}
\end{table}

We evaluate the disagreement between $f_{\mathcal{O}}$ and $f_{\mathcal{C}}$ through the empirical fidelity error $R_{\mathcal{F}}$, the percentage error of the copy over a given set of data points, taking the class predictions of $f_{\mathcal{O}}$ as ground truth~\cite{copying}. To compensate for the potential under-representation of one or more classes in the synthetic dataset, we also compute the \textit{balanced empirical fidelity error}, defined as

$$R_{\mathcal{F}_b}^{\boldsymbol{X}} = \frac{1}{sk} \sum_{j = 1}^{k}\sum_{i = 1}^{s} \mathds{I} [f_\mathcal{O}(\boldsymbol{x}^j_i) =  f_\mathcal{C}(\boldsymbol{x}^j_i)]$$

\noindent
for $k$ the number of classes and $s$ the number of samples per class, so that $\boldsymbol{x}^j_i$ refers to sample $i$ of class $j$. We report metrics averages over 10 repetitions, except for Bayesian sampling, for which we use 5 repetitions. We also provide the execution times of the different methods for different sample sizes. All experiments are carried out in a single m4.16xlarge Amazon EC2 instance with 64 cores, 256 GB of RAM and 40 GB of SSD storage. 

\subsection{Intuition}

\begin{figure}[!t]
    \centering
    \begin{tabular}{c c c c c c}
        \includegraphics[width=0.15\columnwidth]{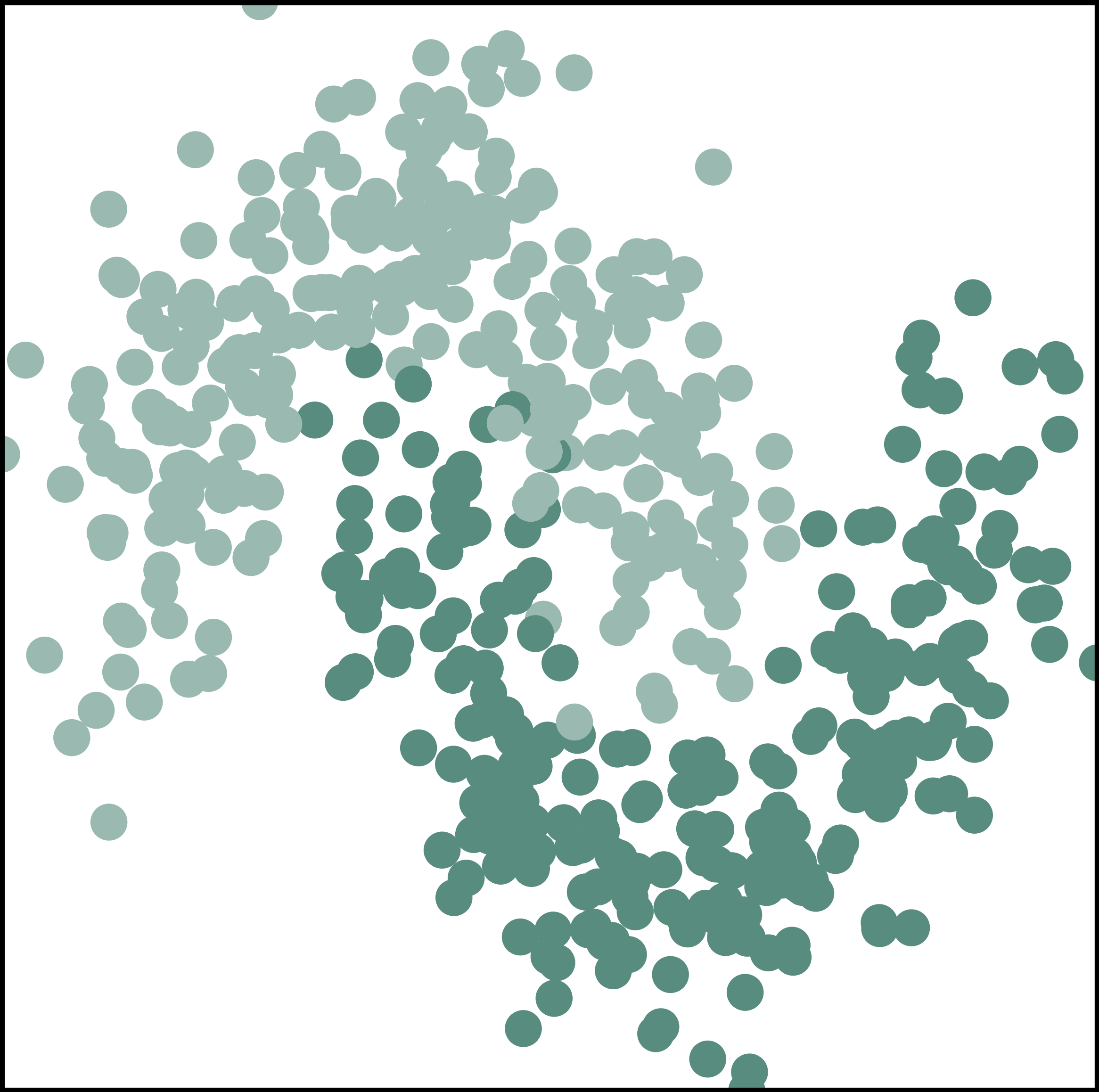} &
        \includegraphics[width=0.15\columnwidth]{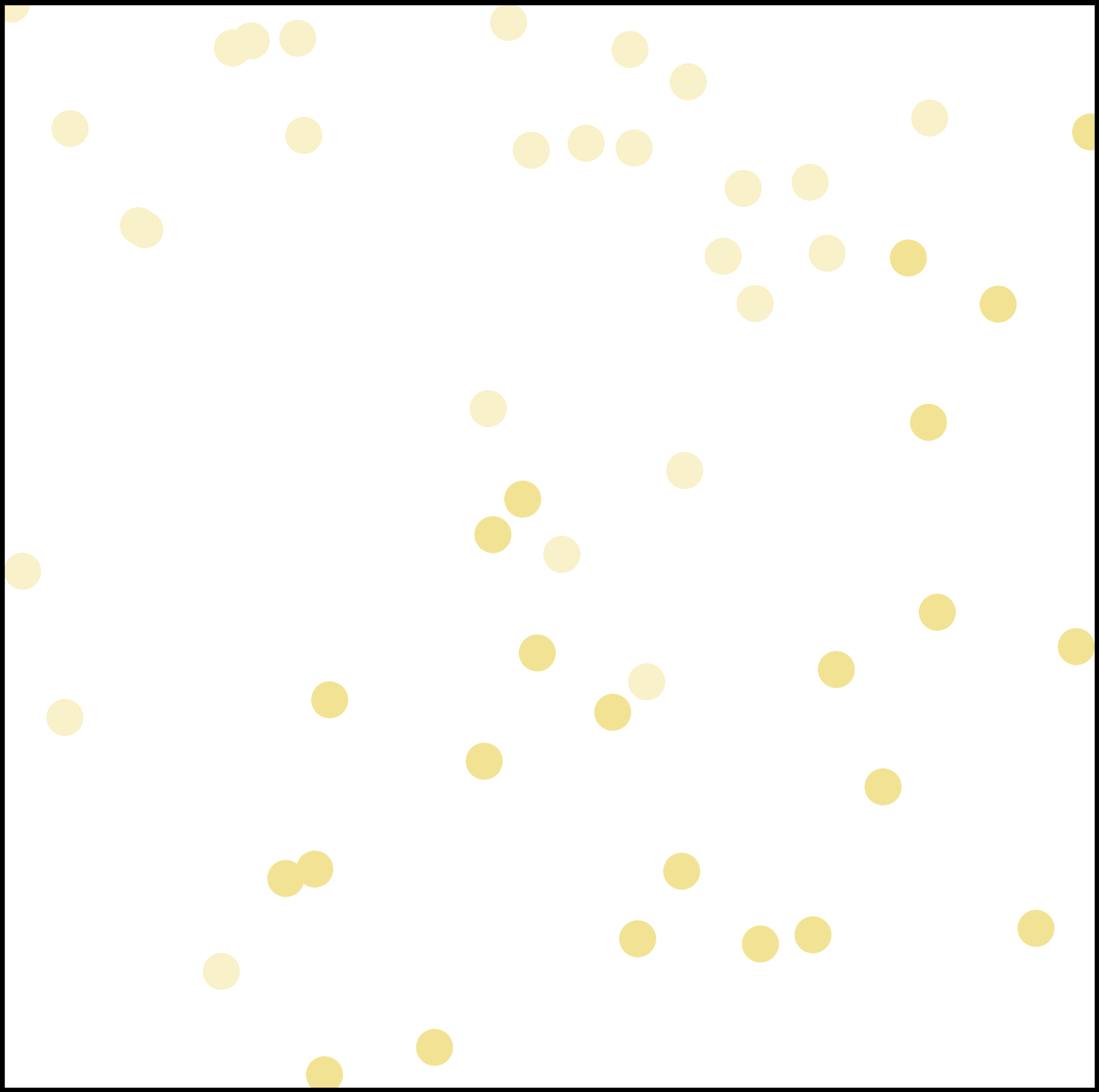} & \includegraphics[width=0.15\columnwidth]{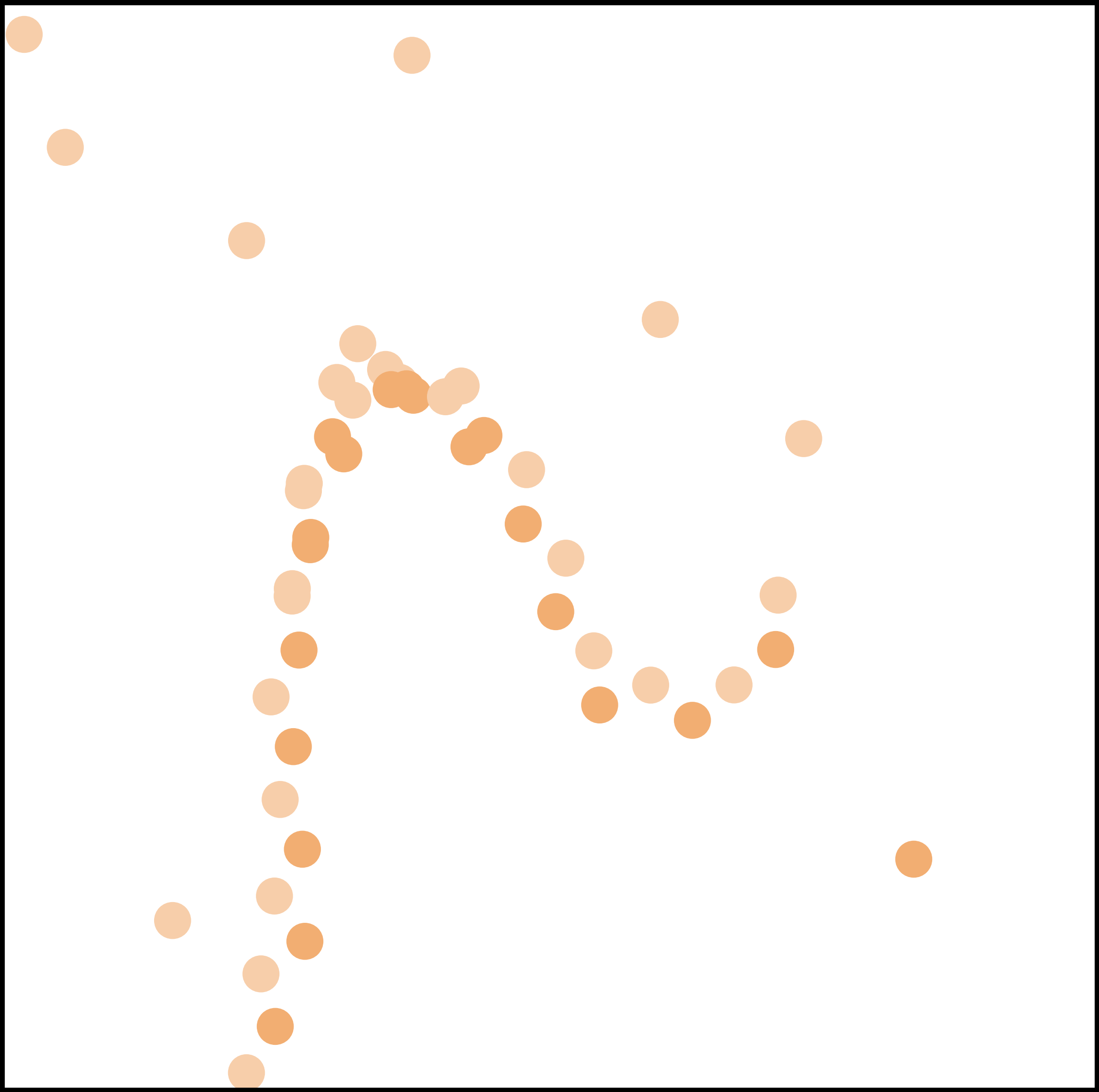} &
        \includegraphics[width=0.15\columnwidth]{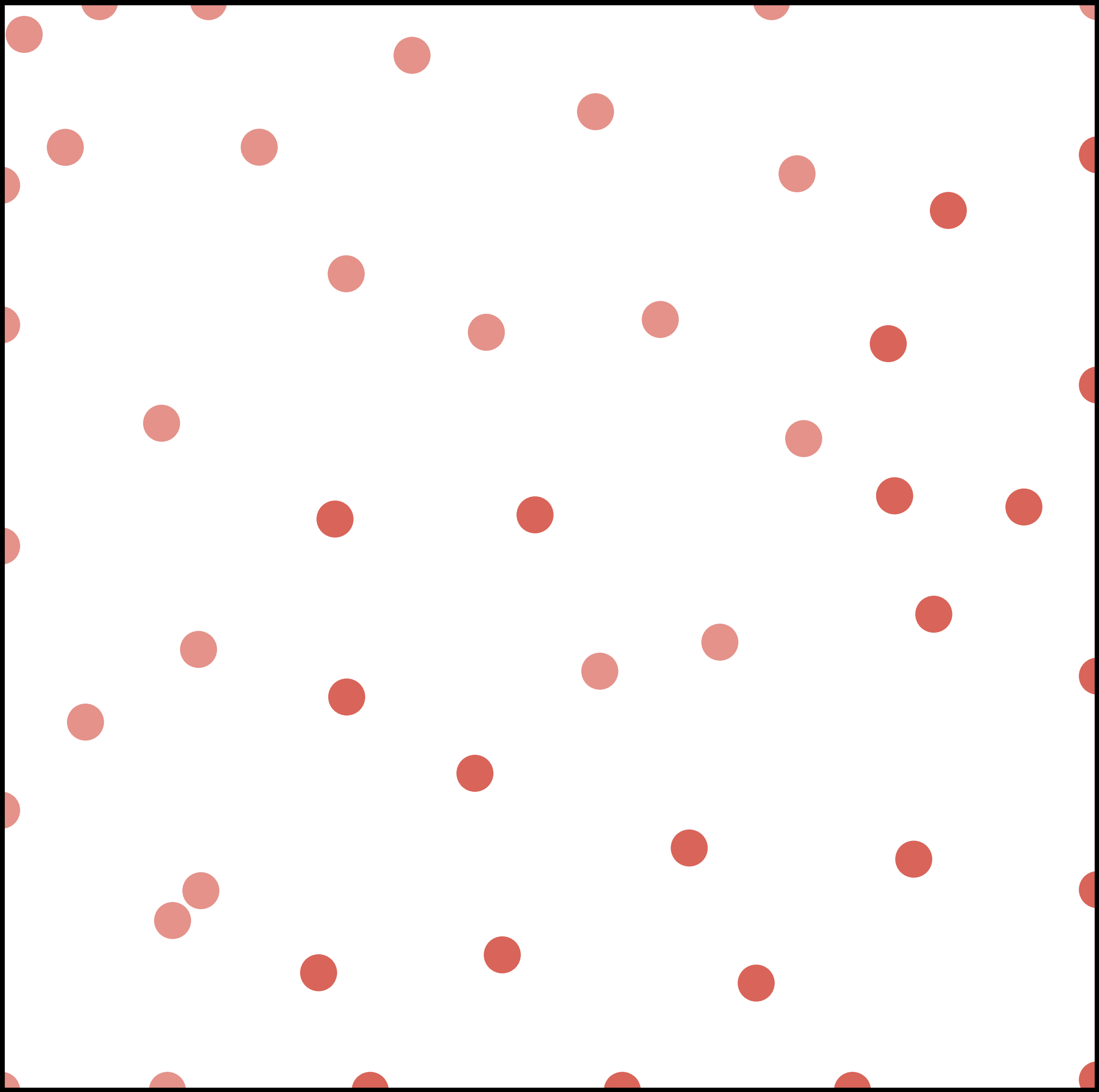} &
        \includegraphics[width=0.15\columnwidth]{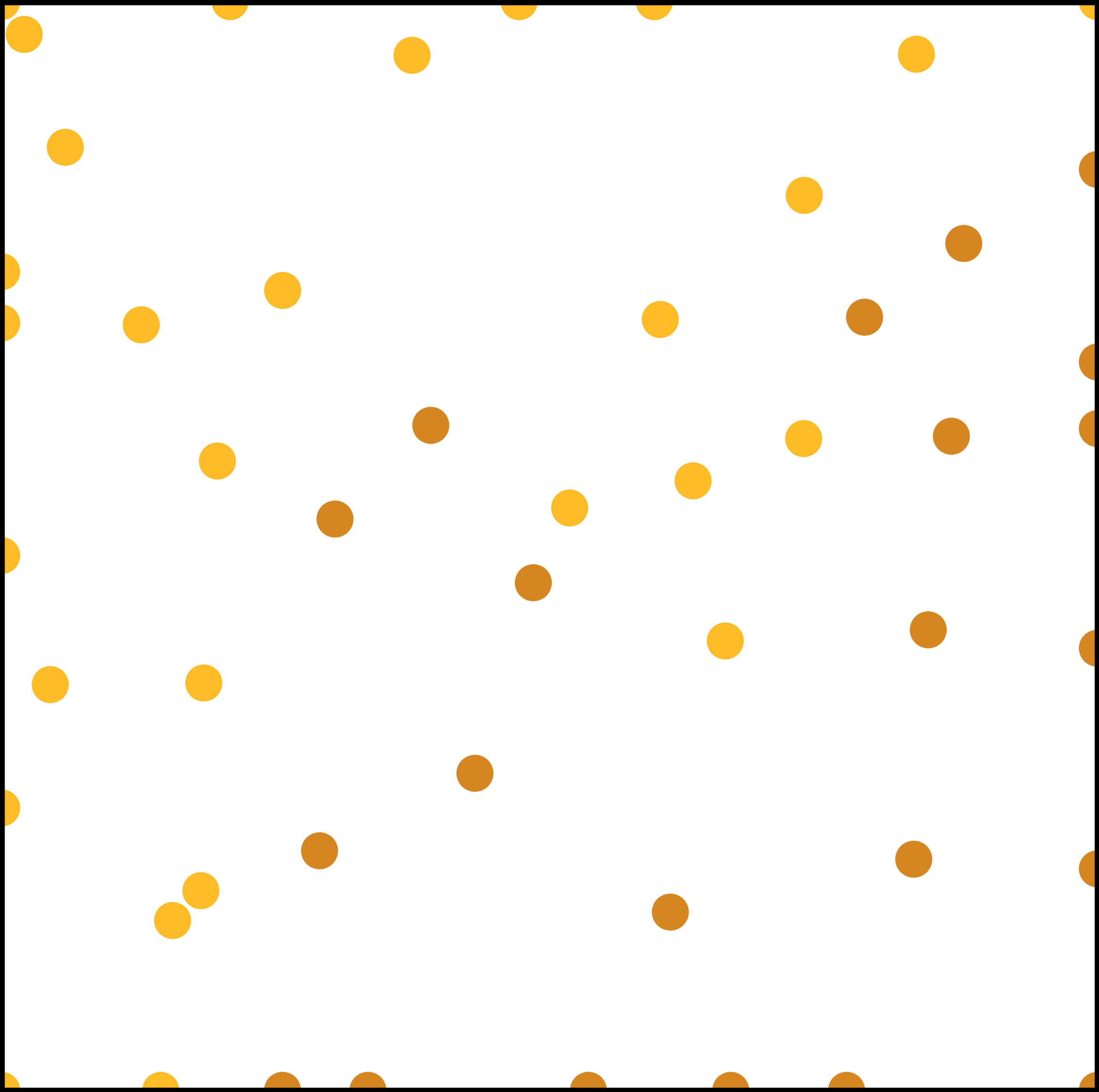} &
        \includegraphics[width=0.15\columnwidth]{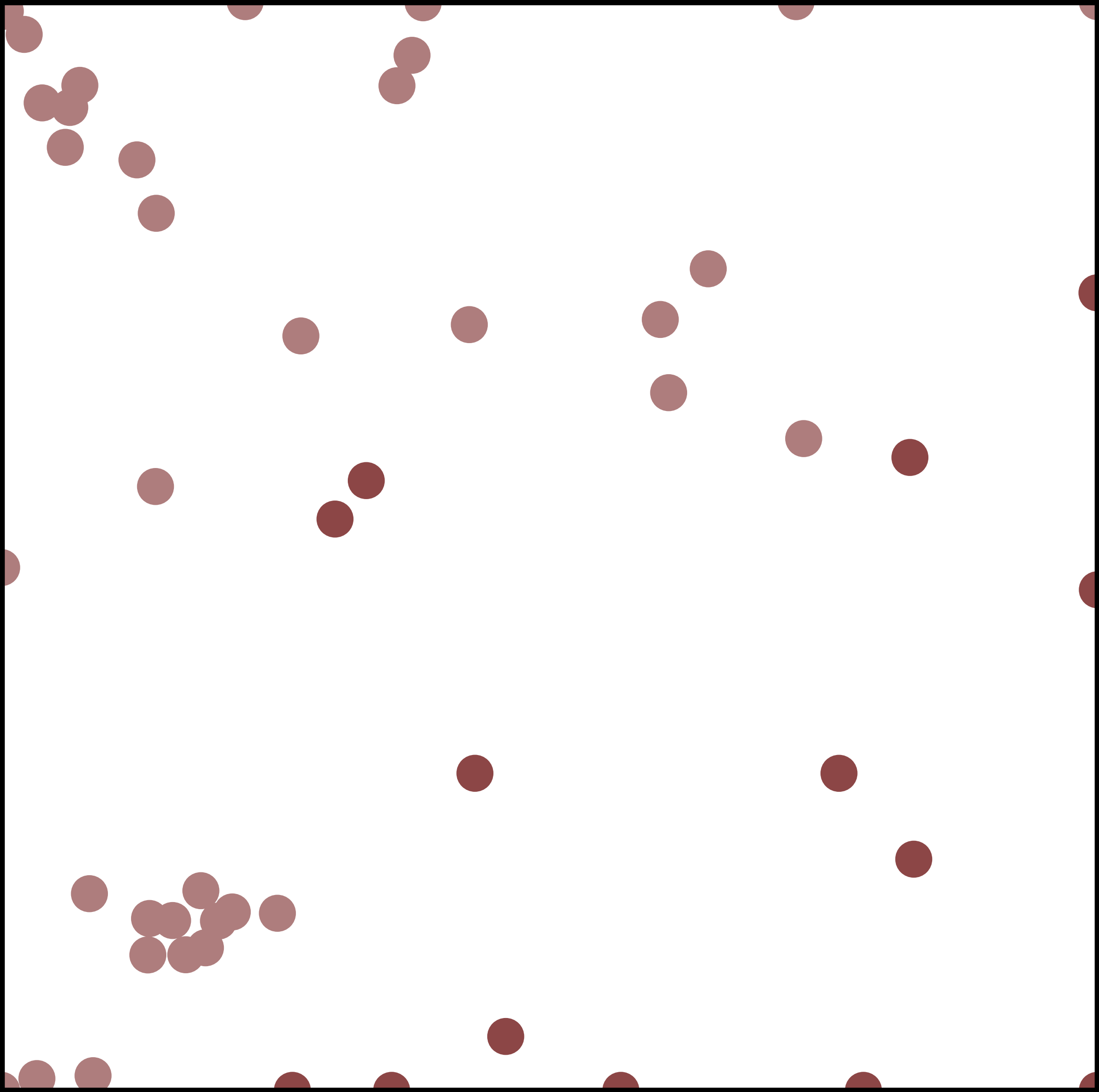}\\
        (a) &
        \includegraphics[width=0.15\columnwidth]{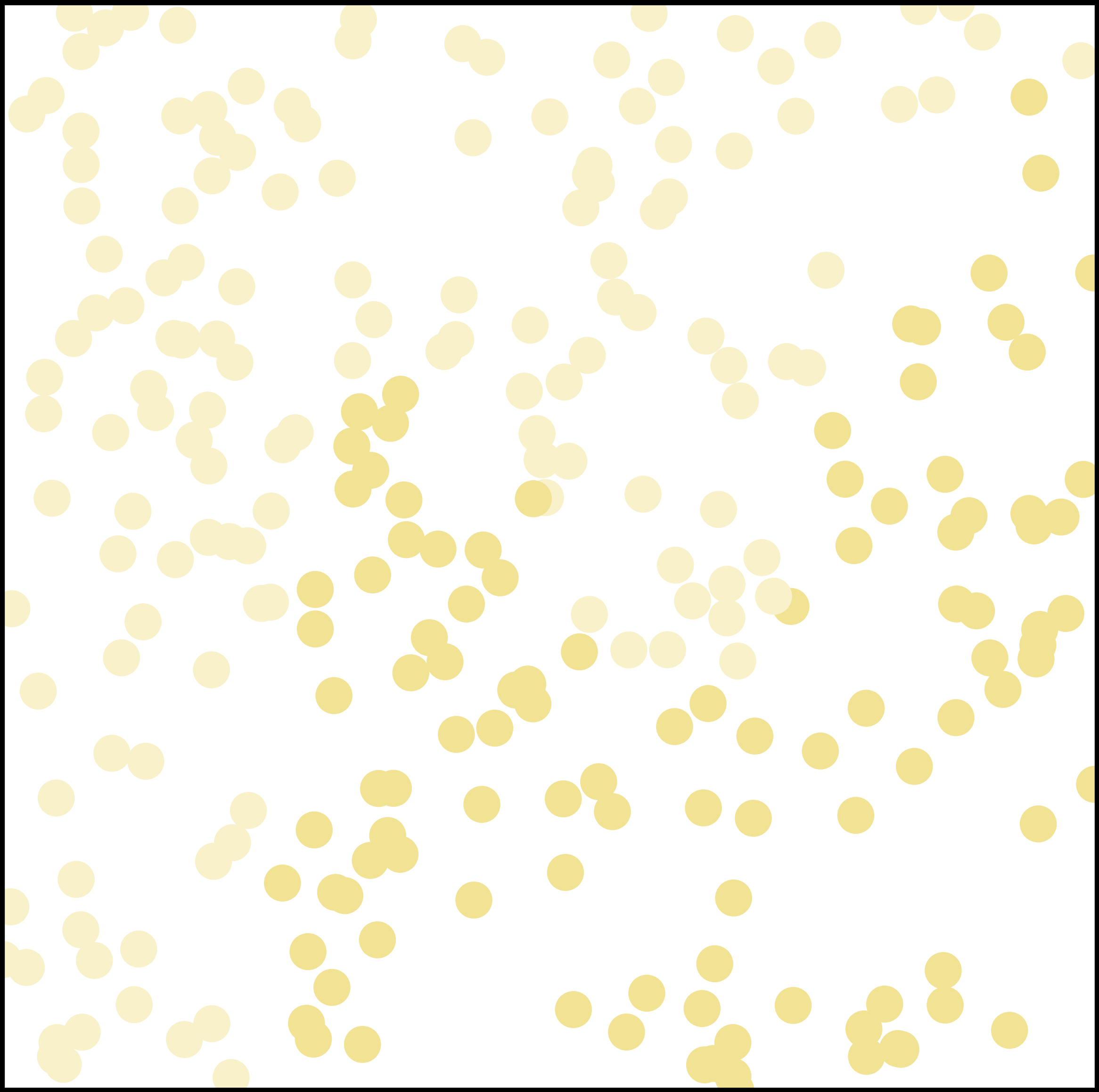} &
        \includegraphics[width=0.15\columnwidth]{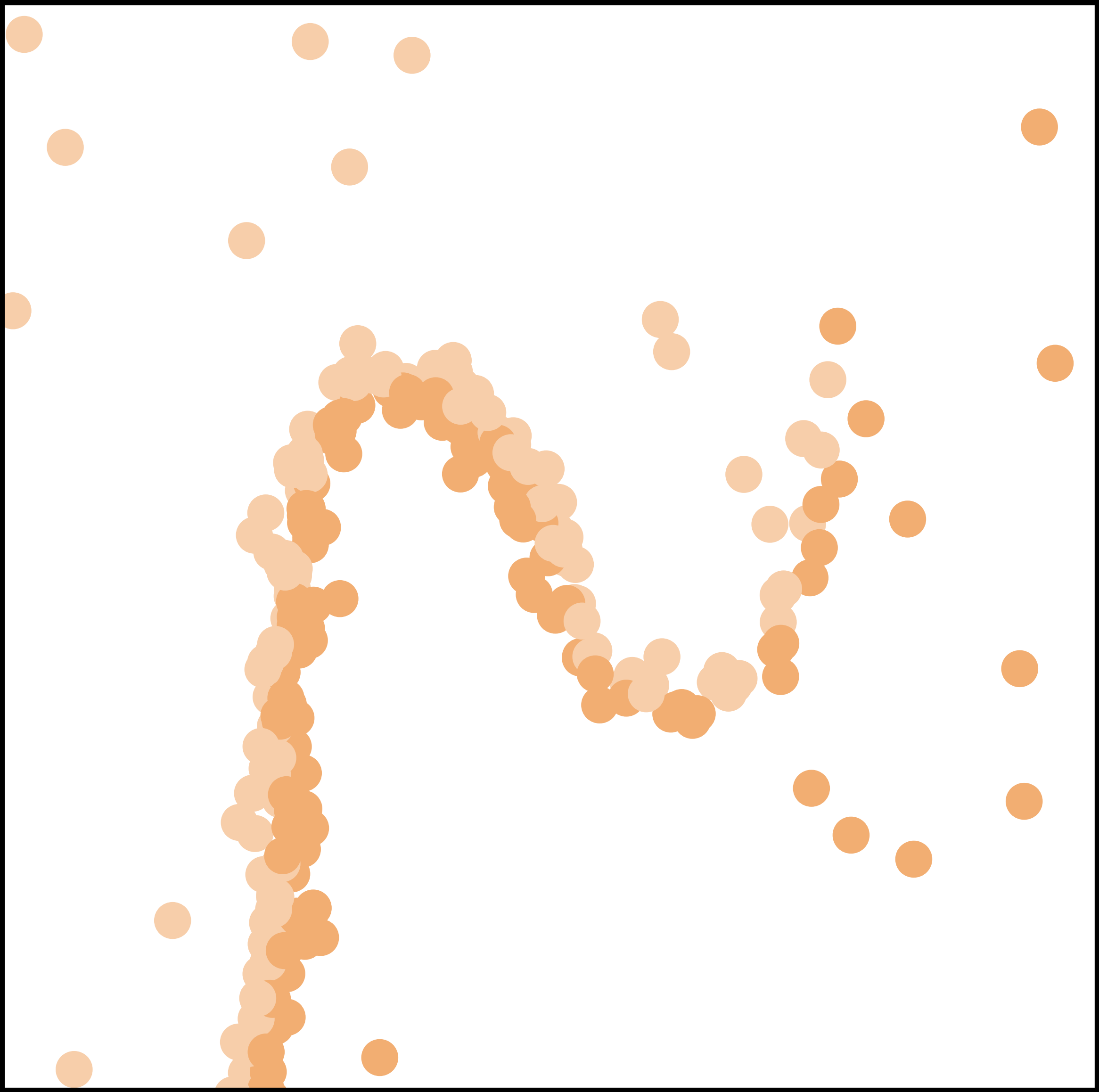} &
        \includegraphics[width=0.15\columnwidth]{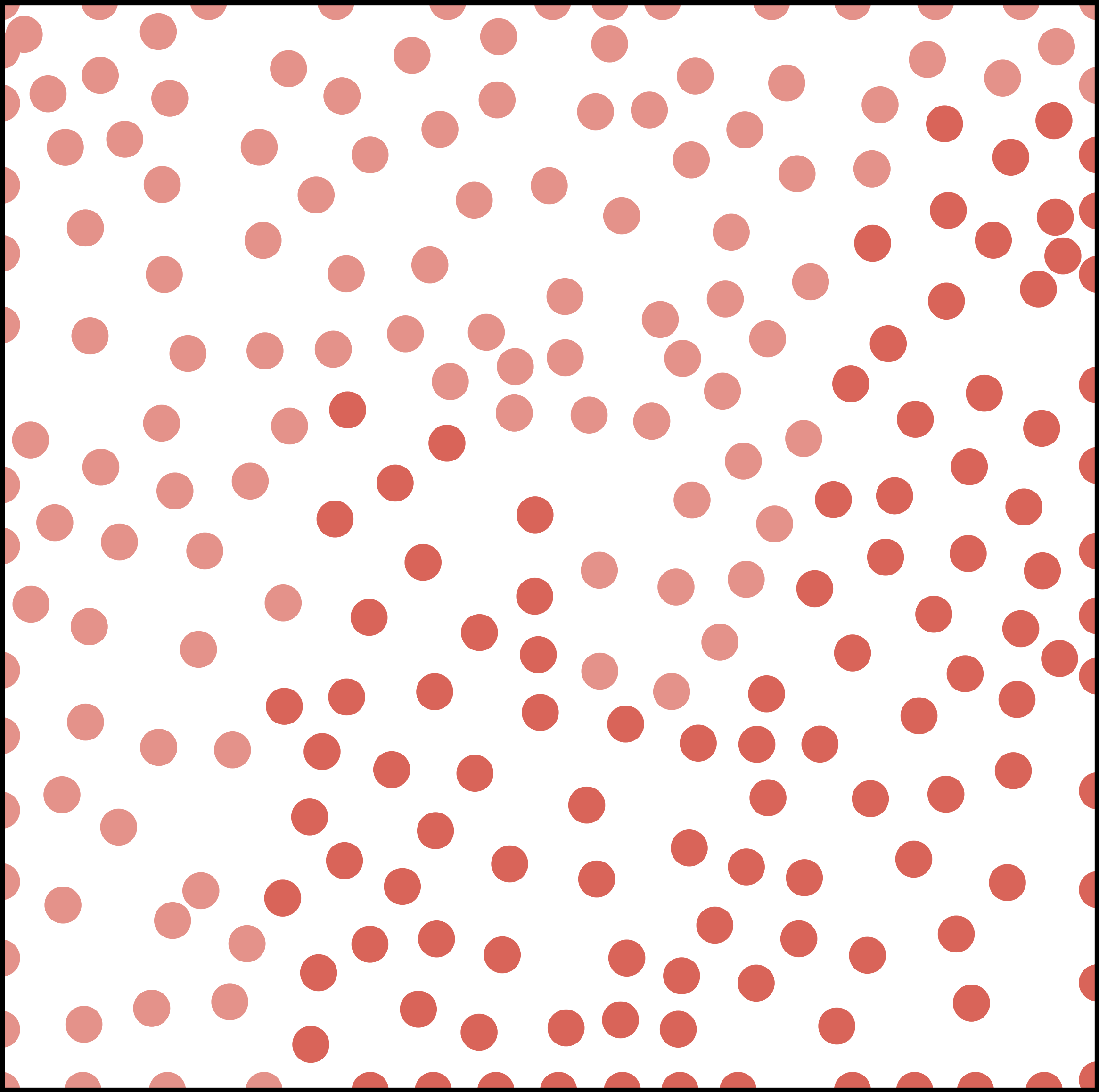} &
        \includegraphics[width=0.15\columnwidth]{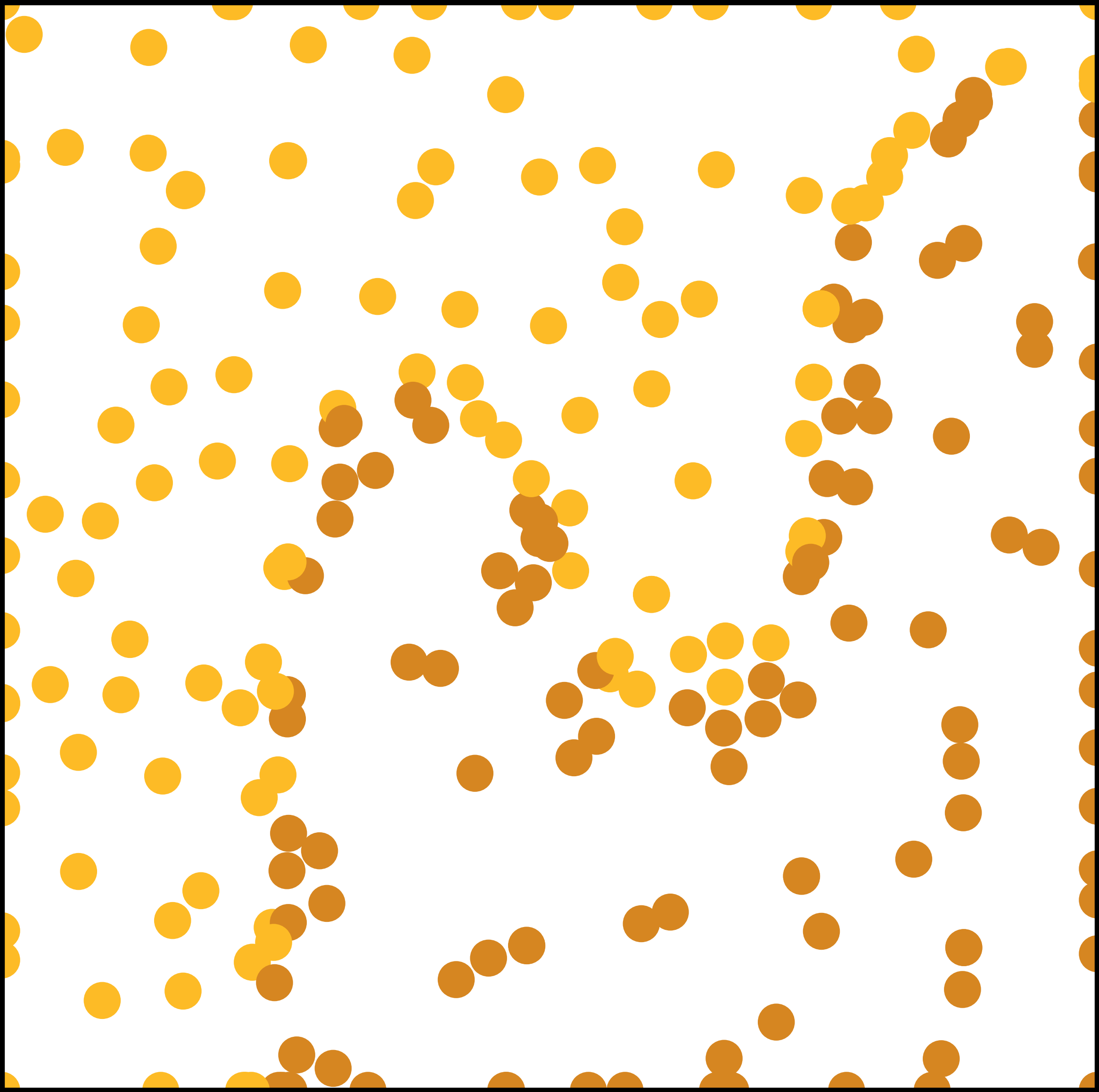} &
        \includegraphics[width=0.15\columnwidth]{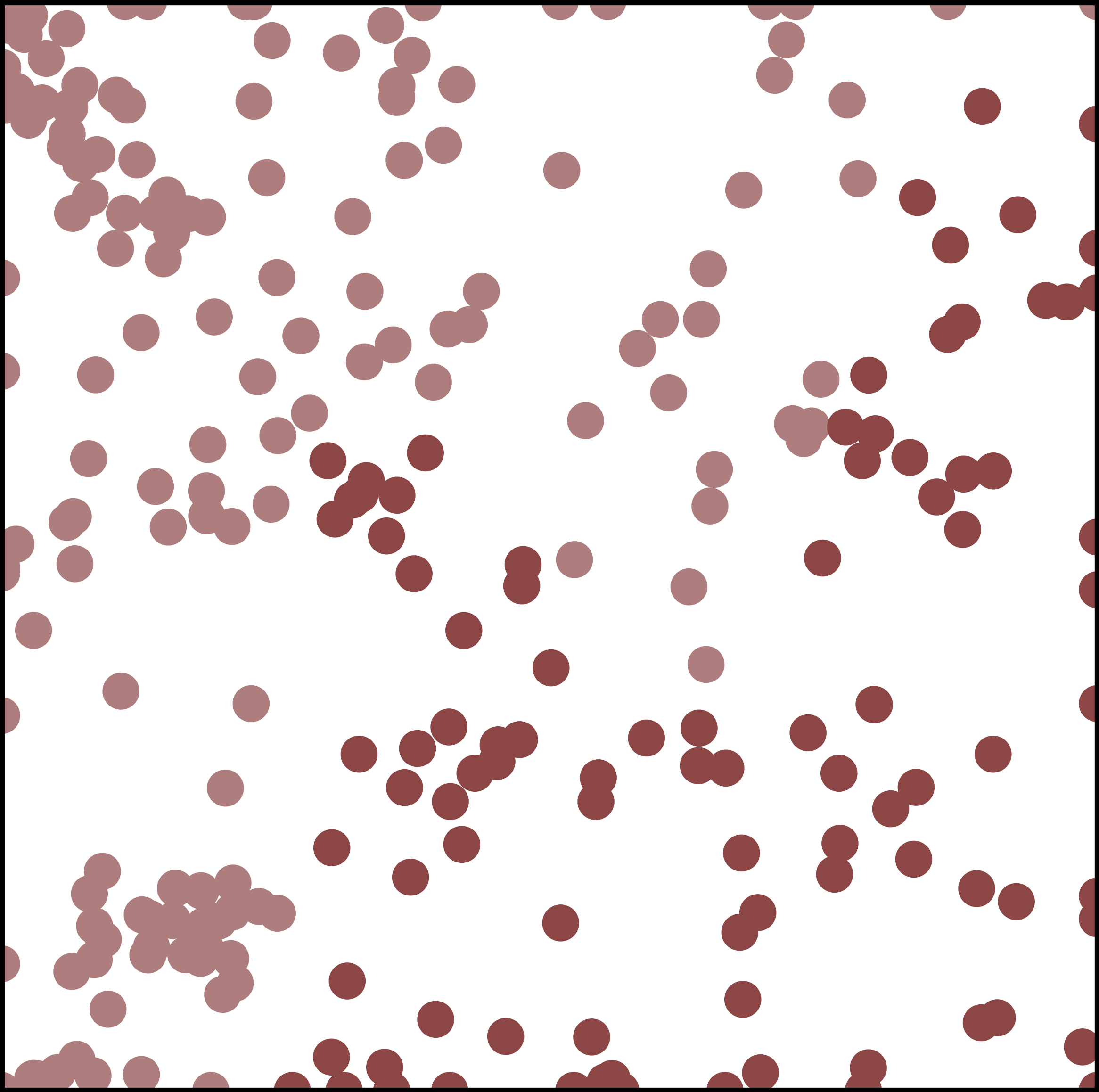} \\ 
        \includegraphics[width=0.15\columnwidth]{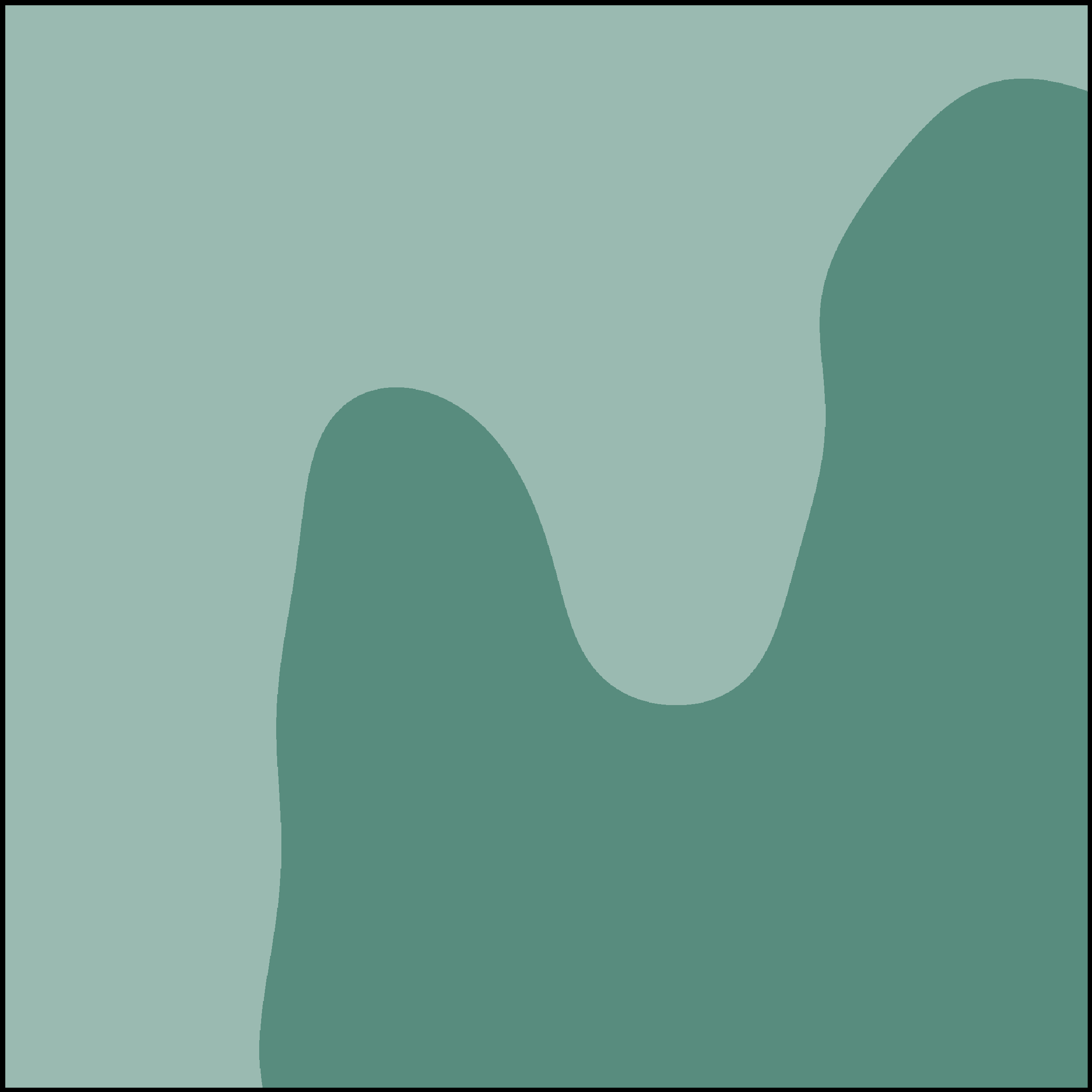} &
        \includegraphics[width=0.15\columnwidth]{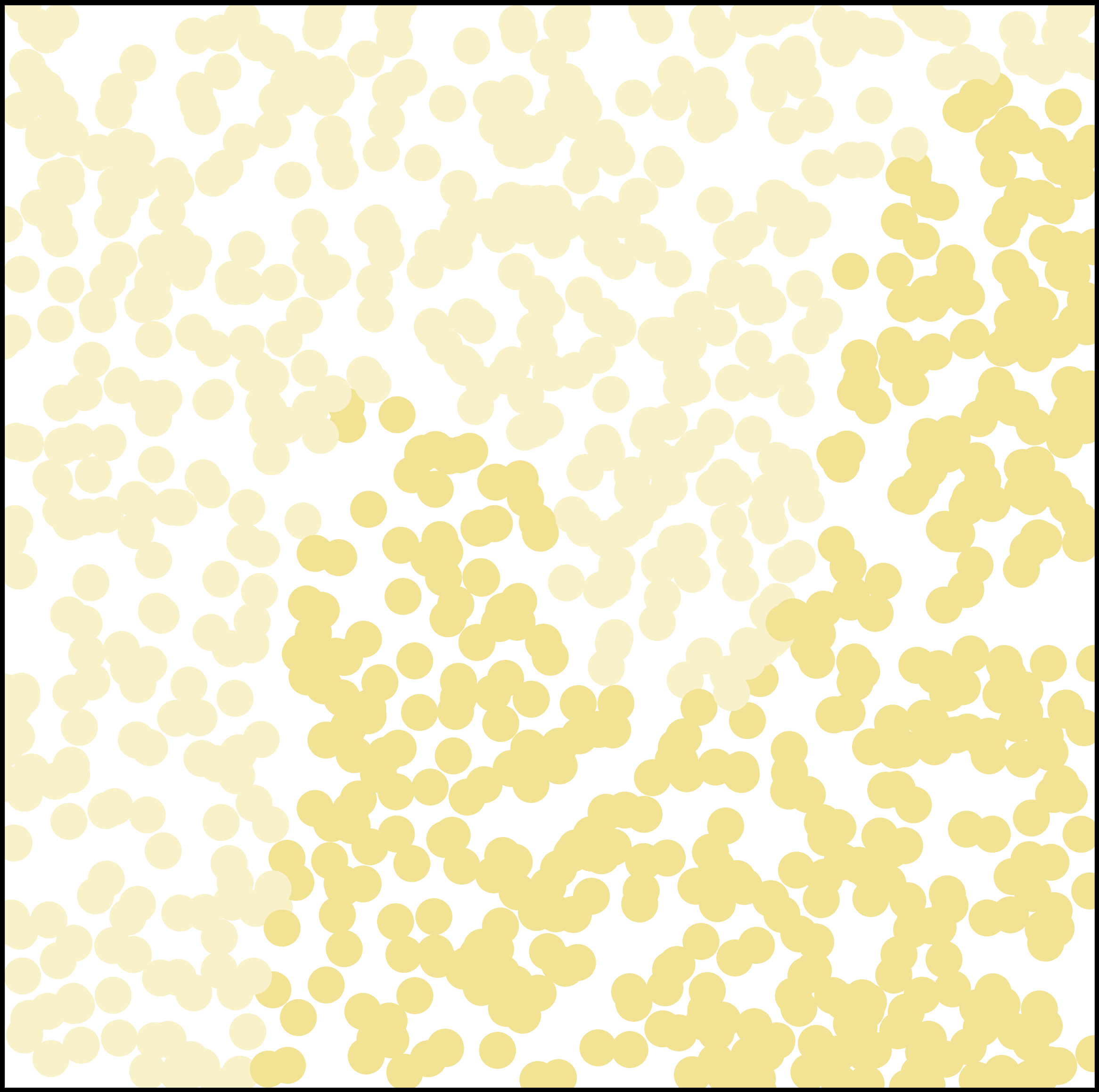} &
        \includegraphics[width=0.15\columnwidth]{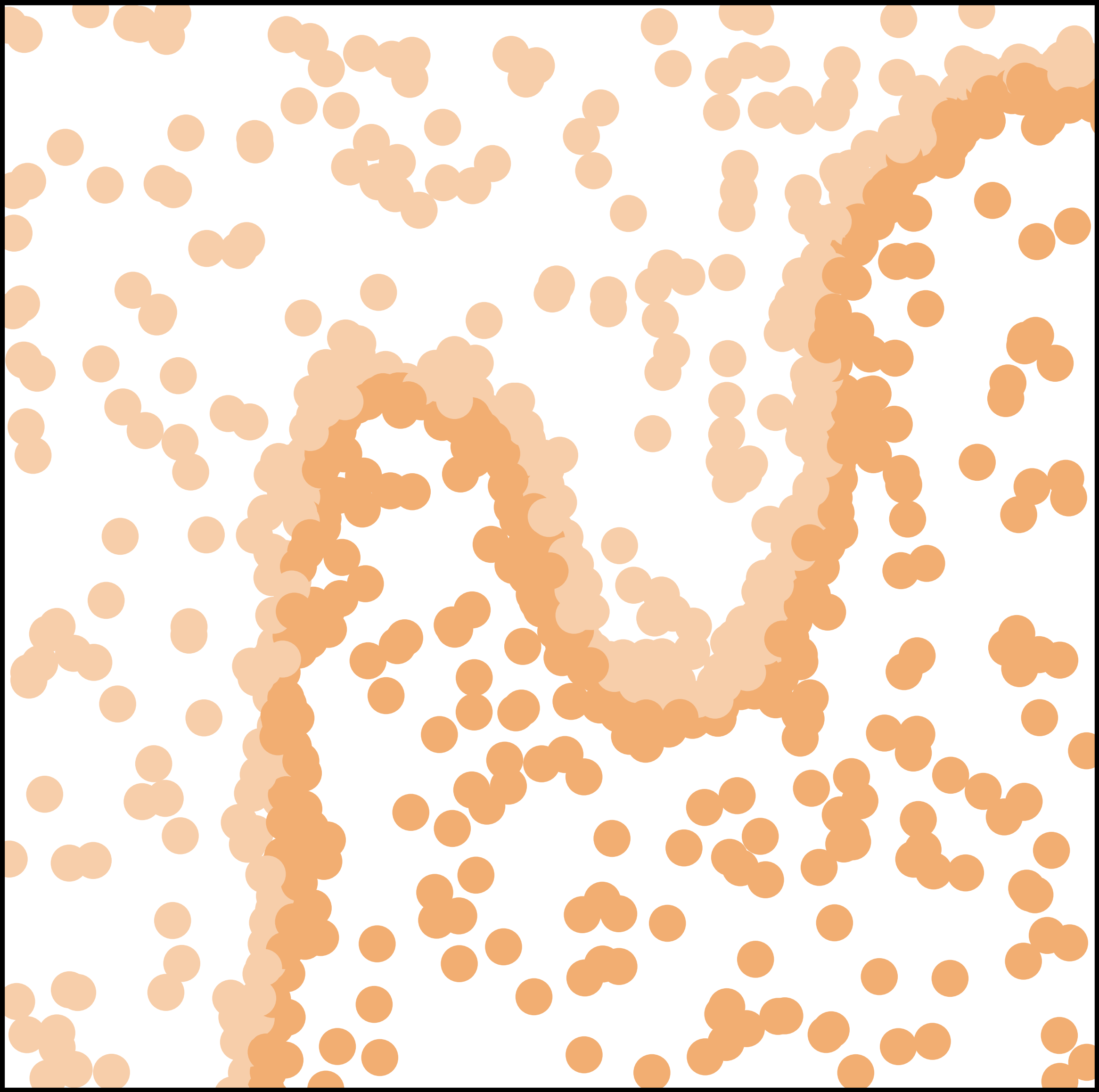} &
        \includegraphics[width=0.15\columnwidth]{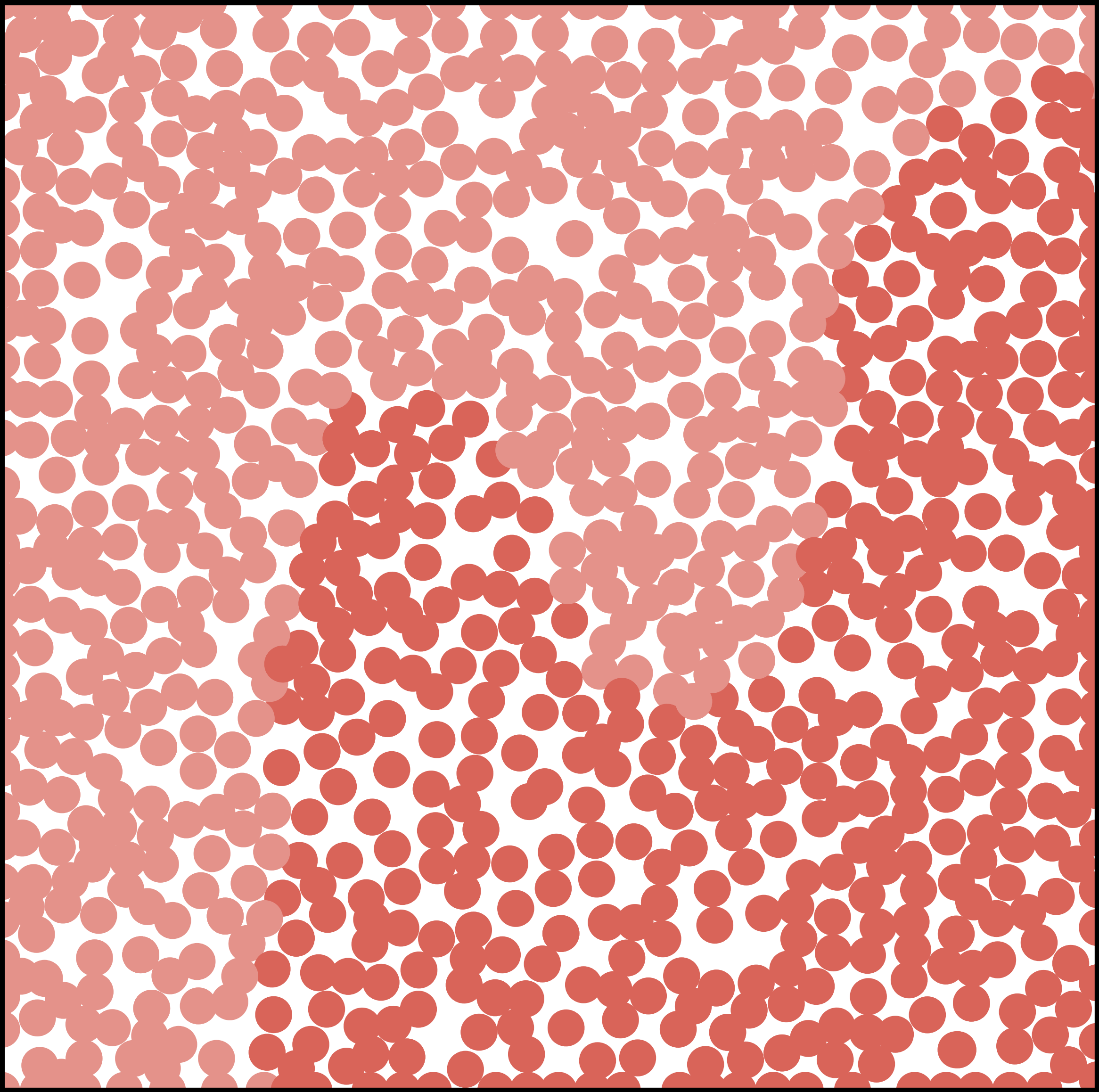} &
        \includegraphics[width=0.15\columnwidth]{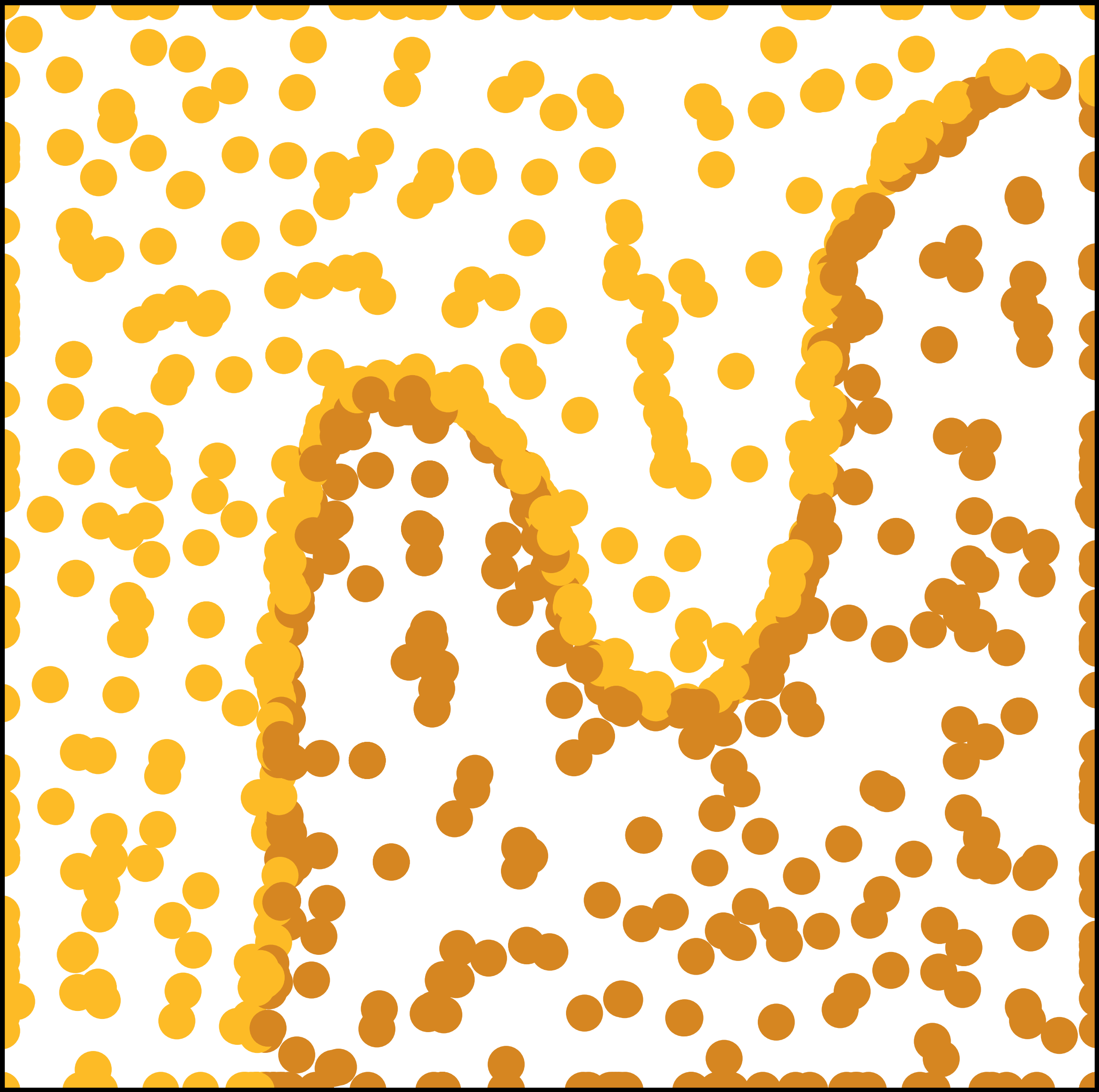} &
        \includegraphics[width=0.15\columnwidth]{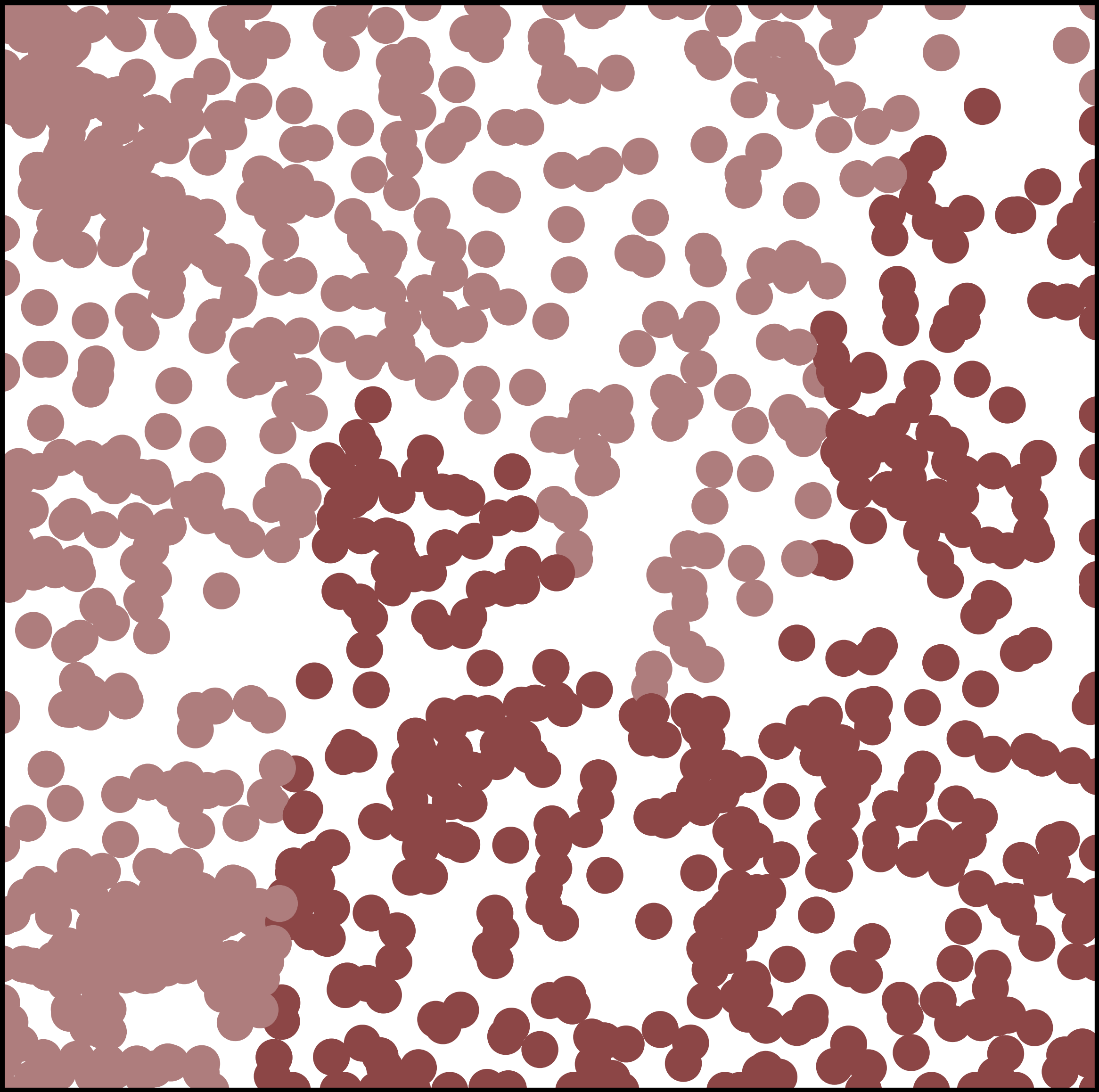} \\
        (b) & (c) & (d) & (e) & (f) & (g) \\
    \end{tabular}
    
\caption{(a) Training dataset and (b) decision boundary learned by an SVM with a radial basis function kernel. Synthetic datasets of sizes 50, 250 and 1000 generated using (c) Random sampling, (d) Boundary sampling, (e) Fast Bayesian sampling, (f) reoptimized Bayesian sampling and (g) adapted Jacobian sampling.}
\label{fig:toy_sampling}

\end{figure}

Before discussing our results we provide an intuition of how the different methods perform on a toy dataset. Examples of synthetic sets for each algorithm are displayed in \figurename \ref{fig:toy_sampling} for different number of samples.

When comparing results for Boundary sampling in \figurename \ref{fig:toy_sampling}(d) to the case of exclusive uniform random sampling in \figurename \ref{fig:toy_sampling}(c), the boundary is sampled with more emphasis, although a large number of samples is required to properly cover this region. Indeed, the main advantage of Random sampling is that it scatters the samples across the space with equal probability. However, this method is oblivious to the structures of interest, \textit{i.e.} it retains no knowledge about the form of the decision boundary for future sampling steps.

Samples generated from Bayesian sampling, in \figurename \ref{fig:toy_sampling}(e), are well distributed, without forming clusters. The importance given to the boundary on the acquisition function is not manifested in a big scale due to the simplifications. If we let the Gaussian process re-optimize several times, it correctly samples the boundary as well as the rest of the space, as shown in \figurename \ref{fig:toy_sampling}(f). Finally, synthetic data points generated with the adapted Jacobian sampling, in \figurename \ref{fig:toy_sampling}(g), form diagonal lines due to the use of the sign of the gradient.

\section{Discussion of results}
\label{sec:discussion}

In what follows we discuss our main experimental results. We first validate the generated reference sample set and then discuss the performance of the different methods, as well as their associated computational cost.

\subsection{Reference set evaluation}

We propose two checks to validate the generated reference sample sets. First, we fit the original architecture to the reference data and compute the corresponding balanced empirical fidelity error, $\mathcal{R}_{\mathcal{F}_b}^{\mathcal{W}}$. As a complementary check, we also evaluate the empirical fidelity error over the original set, $\mathcal{R}_{\mathcal{F}_b}^{\mathcal{D}}$. This value captures how well copies trained on $\mathcal{W}$ generalize.  Table \ref{tab:accuracies} shows the results of the two quality checks. In all cases we observe values close to 0, which we take as an indication that the reference sample sets are a suitable baseline with which to compare our proposed sampling strategies.  We note the exception of the \textit{ilpd-indian-liver} dataset, for which we are not confident enough of our evaluation.

\subsection{Algorithm evaluation}

\begin{figure*}
\centering
    \includegraphics[angle=90,origin=c, scale=0.81]{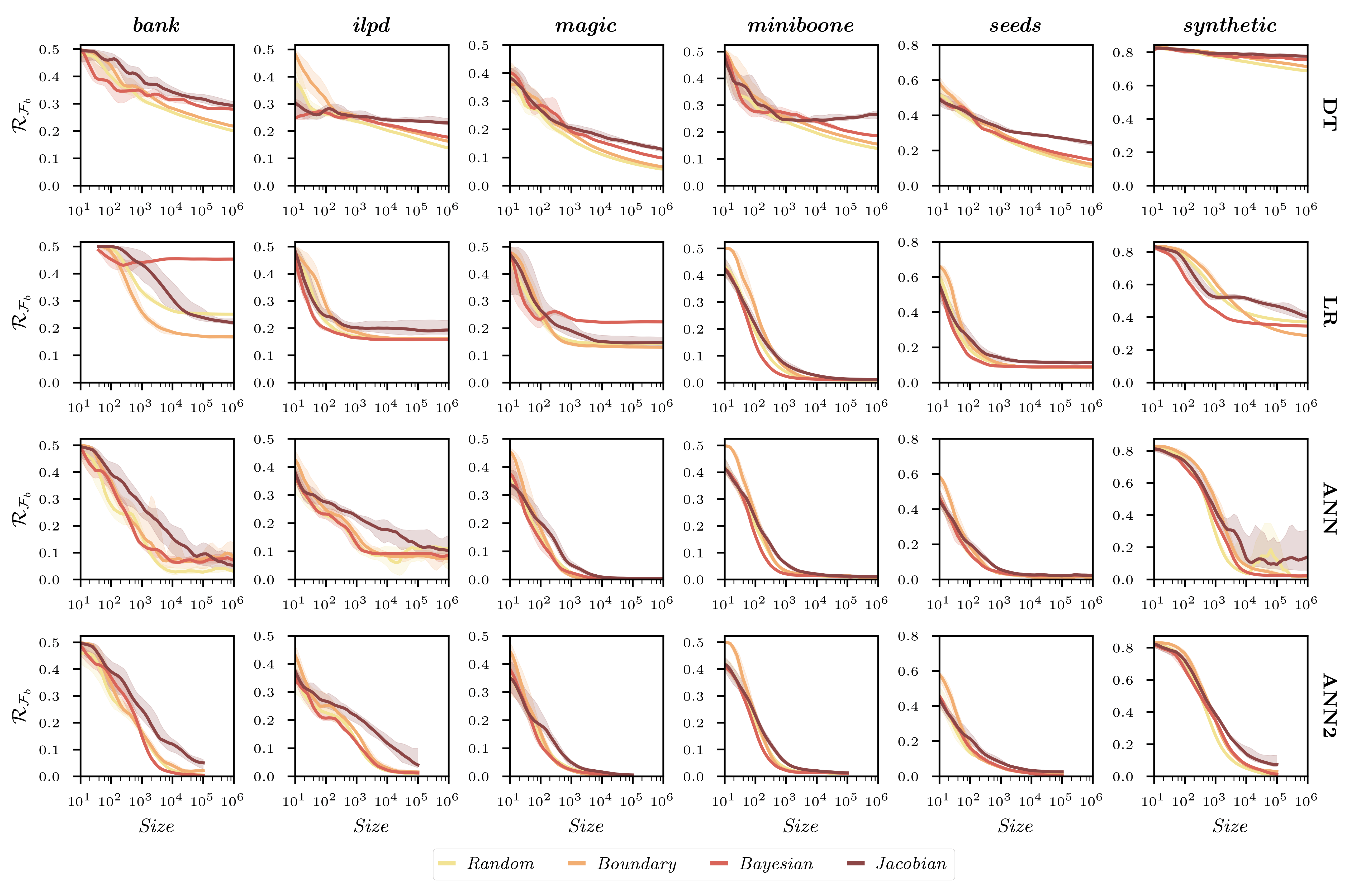}
    \caption{Median and 20-80 percentil band. The similarity axis starts from $1/k$ for $k$ the number of classes: the expected score for a classifier that has not learned anything. For ANN2 models, we only show results for  $10^5$ samples, due to the high training times.}
    \label{fig:graf_acc}
\end{figure*}

In \figurename \ref{fig:graf_acc} we report the balanced empirical fidelity error for the different copy architectures, sampling strategies and datasets, measured on the reference sample sets. Plots show the 20, 50 and 80 percentiles of the multiple realizations.

Boundary sampling performs well for copies based on LR. This technique draws many samples near the boundary, so that the LR has a lot of information to find the optimal decision hyperplane. However, Bayesian sampling displays the fastest growth, \textit{i.e.} performs comparably better with fewer samples. This is because it focuses on globally reducing the uncertainty during the first steps. In the case of LR, it learns fast until it reaches its capacity limit. 

For DT models, Random sampling displays the best behavior, as opposed to algorithms that exploit the boundary. This may be because DTs work well when there is a sample in each region of the space in order to create the leaves. In high dimensionality, the coverage of $\mathcal{X}$ with DTs is costly, which seems to be in accordance with the slowly increasing score obtained for this architecture. 

Copies based on ANN and specially on ANN2 achieve the best scores in general. We highlight the cases of \textit{bank} and \textit{ilpd-indian-liver} datasets for which the simpler ANN performs significantly worst, indicating that the use of matching architectures does not guarantee a good performance. This may happen because the characteristics of the problem change when using the synthetic dataset instead of the training data. This, together with the fact that the original architecture has just enough degrees of freedom to replicate the decision boundary, deters the copy from converging to the same solution.

Table \ref{tab:comp} shows the aggregated comparison among all sampling techniques. In general terms, Jacobian sampling seems to perform the worst. This method generates linear structures which contain a large number of samples. As a result, it has a wide uncertainty band. On the contrary, when a large number of synthetic data points is considered, Random sampling is the algorithm that gathers a greater number of victories. Closely behind, Boundary and Bayesian sampling are both reasonably similar in terms of their averaged performance.

\begin{table}[!t]
\caption{Quality checks for the reference sample sets.}
\label{tab:accuracies}
\renewcommand{\arraystretch}{1.3}
\centering
\begin{tabular}{L{1cm}||c||c||c||c||c||c}
\hline
& \textit{bank} & \textit{ilpd} & \textit{magic} & \textit{miniboone} & \textit{seeds} & \textit{synthetic}\\
\hline
\hline
$\mathcal{R}_{\mathcal{F}_b}^{\mathcal{W}}$ & 0.023 & 0.080 & 0.001 & 0.009 & 0.020 & 0.010\\ 
$\mathcal{R}_{\mathcal{F}_b}^{\mathcal{D}}$ & 0.021 & 0.385 & 0.001 & 0.168 & 0.000 & 0.000\\ 
\hline
\end{tabular}
\end{table}

\subsection{Computational cost}

\begin{figure}[!t]
    \centering
    \includegraphics[width=0.8\columnwidth]{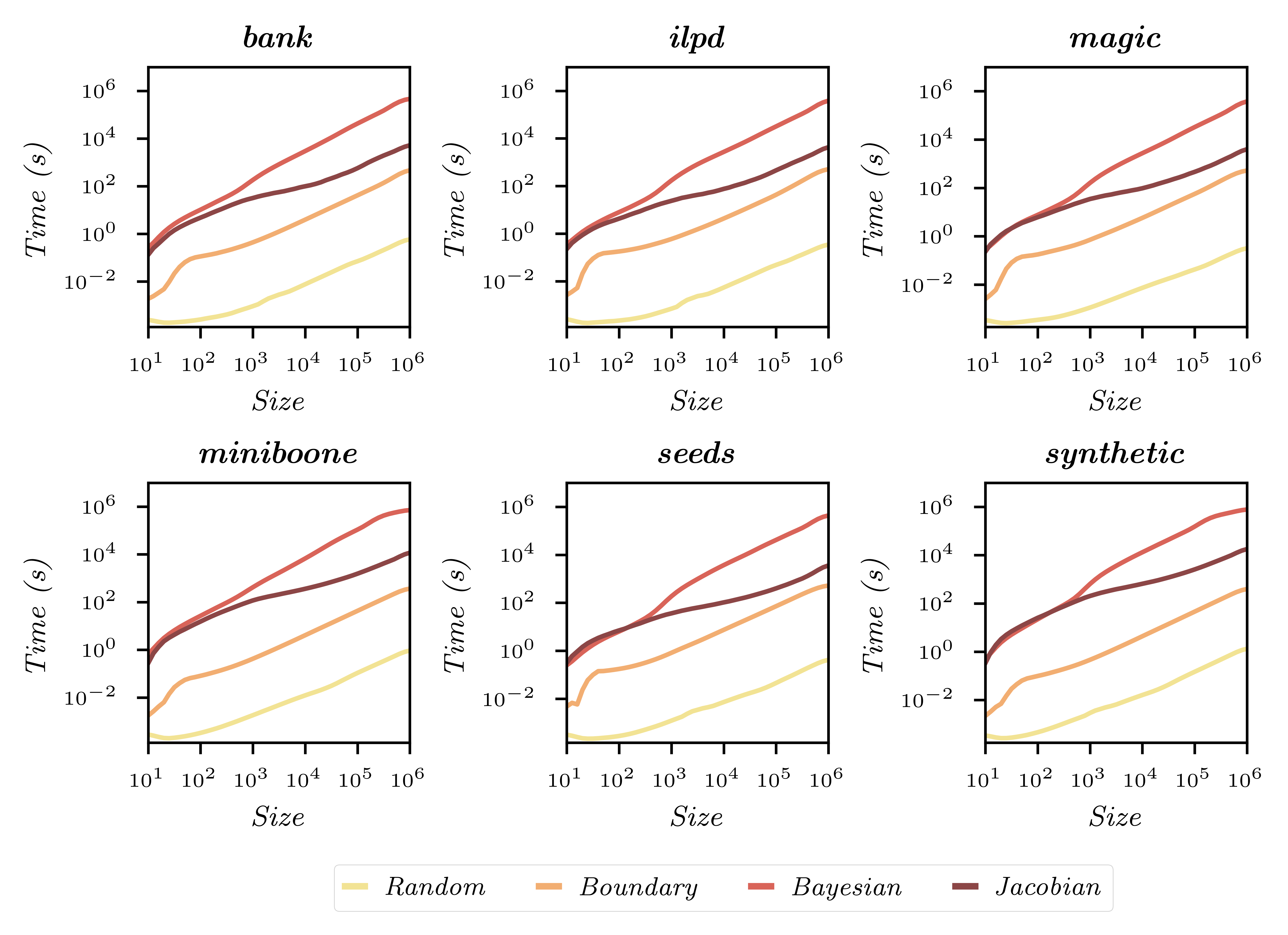}
    \caption{Execution time of the different sampling strategies as a function of dataset size.}
    \label{fig:times}
\end{figure}

In \figurename \ref{fig:times} we compare the algorithms in terms of their computational cost by placing timestamps at different points during the execution for copies based on ANN2, i.e. the worst-case scenario. The remaining copy models yield similar if not better results.  The execution time is asymptotically linear. Bayesian sampling is notably slower than the rest of the methods, while Random sampling is the fastest. Indeed, a great advantage of Random sampling is its simplicity, and consequently its low computational cost. The main drawback of this algorithm is, however, that it samples points with no regards to the form of the decision function or the resulting class distribution. In cases where a large number of dimensions or attributes are involved, Boundary sampling may be a good compromise between time and accuracy. In the absence of any time constrain, however, Bayesian sampling ensures a more reliable exploration of the space.

\begin{table}[!t]
\caption{Comparison of the different methods, shown as victory/tie/loss.}
\label{tab:comp}
\renewcommand{\arraystretch}{1.3}
\centering
\begin{tabular}{l||c||c||c||c}
\hline
& \textit{Random} & \textit{Boundary} & \textit{Bayesian} & \textit{Jacobian} \\ 
\hline
\hline
\textit{Random} & - & 8/13/3 & 10/13/1 & 19/5/0 \\
\textit{Boundary} & 3/13/8 & - & 10/11/3 & 18/5/1\\
\textit{Bayesian} & 1/13/10 & 3/11/10 & - & 16/5/3\\
\textit{Jacobian} & 0/5/19 & 1/5/18 & 3/5/16 & -\\
\hline
\end{tabular}
\end{table}

\section{Conclusions and future work}
\label{sec:conclusions}

In this paper we evaluate different algorithms to sample unknown decision functions. We conclude that Bayesian sampling is the most promising, despite its high computational cost, specially in high dimensional datasets. Strategies that focus on boundaries tend to outperform the rest when building copies with low capacity models, such as LR. Random sampling still exhibits an adequate behavior in general settings, its use is not encouraged when the volume of one or more classes is substantially smaller than that of the rest. 

As a future work we will investigate how to make Bayesian sampling faster and search new sampling methods, such as Determinantal Point Processes (DPP). We would also like to extend this work to regression settings with mixed numerical and categorical data. Finally, more \textit{ad hoc} mechanisms to generate optimal synthetic datasets should move on from the decoupled \textit{single-pass copy} to incorporate a simultaneous optimization of the samples and the copy parameters.

\bibliographystyle{splncs04}

\end{document}